\documentclass[manuscript=article]{achemso}

\usepackage[version=3]{mhchem} 

\usepackage{pifont} 
\newcommand{\cmark}{\ding{51}}
\newcommand{\xmark}{\ding{55}}

\usepackage{multirow}
\usepackage{booktabs} 

\usepackage[T1]{fontenc} 
\newcommand{\breaktt}[1]{{\ttfamily\hyphenchar\font=`\- #1}}

\usepackage{hyperref}
\hypersetup{
    colorlinks=true,
    linkcolor=black, 
    citecolor=black, 
    filecolor=black, 
    urlcolor=blue,   
}

\author{Th\'{e}o A. Neukomm}
\email{theo.neukomm@epfl.ch}

\author{Zlatko Jon\v{c}ev}

\author{Philippe Schwaller}
\email{philippe.schwaller@epfl.ch}

\affiliation[LIAC EPFL]
{Laboratory of Artificial Chemical Intelligence
(LIAC), Institut des Sciences et Ingénierie Chimiques, Ecole
Polytechnique Fédérale de Lausanne (EPFL), Lausanne 1015,
Switzerland}

\title[MechSMILES and ChRIMP]
  {Teaching Language Models Mechanistic Explainability Through MechSMILES}

\begin{document}

\begin{abstract}
Chemical reaction mechanisms are the foundation of how chemists evaluate reactivity and feasibility, yet current Computer-Assisted Synthesis Planning (CASP) systems operate without this mechanistic reasoning. We introduce a computational framework that teaches language models to predict reaction mechanisms through arrow-pushing formalism, a century-old notation that tracks electron flow while enforcing conservation of mass and charge. This mechanistic understanding enables three capabilities that are difficult or impossible with current methods: post-hoc validation of CASP proposals by reconstructing physically plausible electron pathways, holistic atom-to-atom mapping that tracks all atoms including hydrogens, and extraction of catalyst-aware reaction templates that distinguish recycled catalysts from spectator species. Central to our approach is MechSMILES, a compact textual format encoding molecular structure and electron flow through three arrow types, designed within a Python-based environment that enforces conservation laws and eliminates the possibility of atom hallucination. We trained and benchmarked models on four mechanism prediction tasks of increasing complexity using the main mechanistic datasets in the literature. On our most challenging task, predicting complete mechanisms given only reactants, conditions, and the desired product, our models achieve 93.2\% and 73.3\% pathway retrieval on the FlowER and mech-USPTO-31k datasets respectively, with top-3 retrieval reaching 97.6\% and 86.5\%. Furthermore, the framework rapidly learns new reaction classes, with strong mechanistic predictions for ozonolysis and Suzuki cross-coupling emerging from as few as 40 training examples each. By grounding predictions in physically meaningful electron movements, this work provides an architecture-agnostic, open-source foundation for more explainable and chemically valid computational synthesis planning.
\end{abstract}

\section{Introduction}
When evaluating a proposed synthetic step, chemists instinctively reason through its mechanism: they trace electron flow, check whether intermediates are accessible. This arrow-pushing formalism, developed over a century ago, remains one of the main tools by which experimental chemists judge reaction feasibility. It serves as a useful level of abstraction, casting complex quantum phenomena as intuitive electron movements and providing a physically grounded representation that enforces conservation of mass and charge by construction. Yet current Computer-Assisted Synthesis Planning (CASP) systems operate entirely without this reasoning. Recent advances have produced commercial and open-source systems capable of proposing synthetic routes  \citep{synthia, reaxys,segler2018planning, schwaller2020predicting, genheden2020aizynthfinder,schwaller2022machine,saigiridharan2024aizynthfinder, tu2025askcos,bran2025chemical}, but these systems predominantly rely on a retrosynthetic approach, starting from the target molecule and working backwards to reach a set of starting materials. Operating in such a way, most of them suffer from two key limitations. First, they often suggest reactions that are formally valid as graph transformations but chemically implausible due to high-energy intermediates or forbidden electron movements. Second, they lack explainability, as their mechanistic reasoning behind proposed transformations remains opaque.

Various approaches have attempted to address feasibility through scores \citep{ertl2009sascore, coley2018scscore, neeser2024fsscore}, or via the crafting of elaborated expert-curated templates \citep{szymkuc2016computer, synthia, patel2020savi}. Other works have classified single step reactions into reactions classes\citep{kraut2013algorithm, schneider2015development, baylon2019enhancing, schwaller2021rxnfp}, or detected the chemical context of the transformations\citep{dobbelaere2024rxninsight}, increasing the granularity of the analysis. However, these methods deal with net transformations and do not capture the underlying mechanistic logic that experimental chemists use to evaluate reactions. 
We address these limitations by teaching language models to reason about reactions through arrow pushing mechanisms. We developed a computational environment (available as a Python package) that encodes the rules of the arrow pushing formalism and trained models on four mechanism prediction tasks of increasing complexity. Our most challenging task mirrors the problem human chemists face: predicting the complete mechanism given only the available reactant species and main product, without knowing stoichiometry or by-products. 

To enable this approach, we developed MechSMILES, a compact and unambiguous textual format for representing mechanistic steps. MechSMILES concatenates a mapped standard Simplified Molecular Input Line Entry System (SMILES) \citep{weininger1988smiles} string with a suffix encoding electron movements through three arrow types: attacks, ionizations, and bond attacks. This format is both human-readable and suitable for training language-based models, bridging the gap between chemical notation and machine learning.

Automated mechanism prediction from reactions using arrow pushing has been addressed via expert curated rule-based system \citep{jorgensen1990cameo, hollering2000simulation, chen2009no, kayala2011learning, klucznik2024computational}, with an increasing shift towards more data-driven techniques in \citet{bradshaw2019generative}, \citet{joung2024reproducing}, and in \citet{miller2025mechanism}. Most recently, \citet{chen2025predicting} and \citet{joung2025electron} trained models to predict multi-step synthesis one elementary step at a time, functioning as mechanistic “reasoning traces” that generate mechanisms before predicting products. Our work is complementary: rather than generating products through mechanisms, we operate as a “post-hoc rationalizer” that validates proposed reactions by reconstructing their mechanisms. This distinction is important for practical integration with existing CASP workflows. State-of-the-art systems propose single-step transformations without mechanistic decomposition; our models can validate these proposals by searching for physically plausible electron pathways, effectively serving as a chemical feasibility filter for any CASP system, whether template-based or template-free.

Beyond validation, mechanism prediction unlocks additional capabilities that are difficult or impossible with current automated methods. By tracking electron flow, we enable holistic atom-to-atom mapping that includes all hydrogen atoms, critical for reactions where hydrogen transfer determines the outcome. We can also extract catalyst-aware reaction templates that distinguish catalysts recycled during the mechanism from inert spectator species, a distinction invisible to traditional template extraction methods that only compare initial and final states.

\begin{figure}[ht!]
\centering
\includegraphics[width=0.9\linewidth]{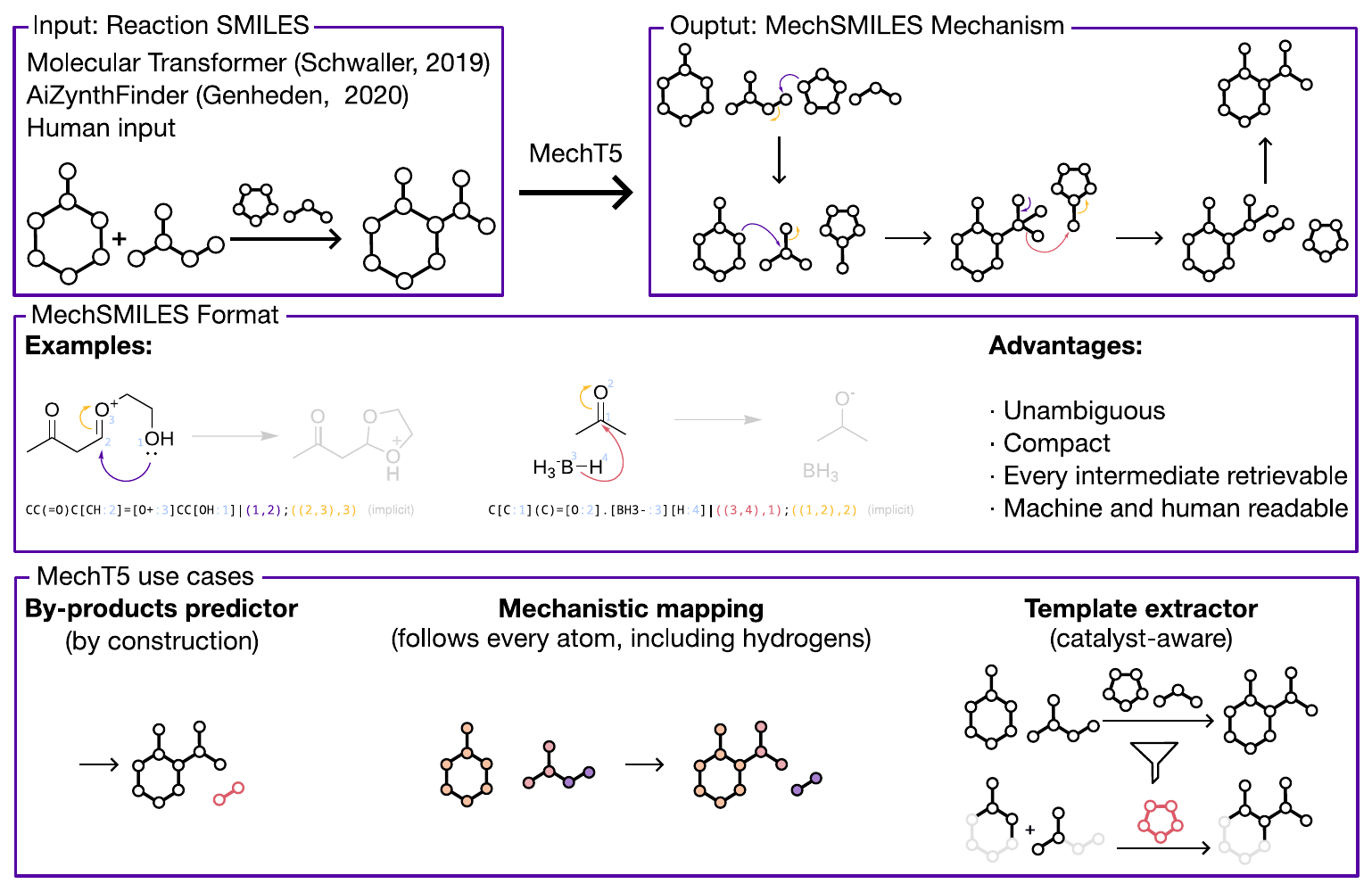}
\caption{Overview of this work. The models trained (illustrated here as MechT5) are taking chemical reactions as input, and will try to produce the optimal mechanism from reactants to products through arrow-pushing. To do so, the models use our MechSMILES format for arrow-pushing moves, with several advantages over existing methods. Once the mechanism is known, several downstream insights become available, such as by-product prediction, mechanistic mapping, or catalyst-aware template extraction.}
\label{fig: overview}
\end{figure}

\section{Methods}
\subsection{MechSMILES format}\label{sec: mechsmiles}

To represent mechanistic steps in a form suitable for both human interpretation and language model training, we developed MechSMILES, a compact and unambiguous textual format for arrow pushing. A MechSMILES string is formed by concatenating a minimally mapped SMILES and its electron flow, composed of three arrow types. (1) The attack, written \texttt{(a, b)}, describes the attack of a lone pair of atom a on atom b, increasing their bond-degree by one; (2) the ionization, written \texttt{((a, b), b)}, describes the heterolytic cleavage of the bond a–b, positively ionizing a and negatively ionizing b, while decreasing the order of their bond by one; (3) The bond attack, written \texttt{((a, b), c)}, describes the attack of the bond a-b on a third atom c through b. This move simultaneously decreases the degree of the bond a-b by one, and increases the degree of the bond b-c by one. Two example MechSMILES are depicted as examples in Figure \ref{fig: overview}.

MechSMILES offers three key advantages for conveying a mechanistic action (Table \ref{tab: advantages_of_mechsmiles}).
First, explicit hydrogen representation gives the models the ability to select individual hydrogen atoms, which is essential for several mechanisms. For example, in E2 eliminations that deviate from Zaitsev’s rule due to steric constraints, the identity of the eliminated hydrogen could determine the outcome; MechSMILES can express this distinction directly.
Second, its step-wise flexibility allows the model to include only the molecules interacting at each step, implicitly capturing the sequential order of reagent addition. 
Third, because MechSMILES encodes only reactants and arrows (the product is computed by our environment), the format achieves a character efficiency 44.6\% greater than the next densest alternative (Table \ref{tab: advantages_of_mechsmiles}, Figure \ref{fig: char_efficiency_comparison}), reducing cost at both training and inference without sacrificing readability.\\

Several alternative formats for encoding mechanistic transformations have been reported in the literature, each making different design trade-offs. Some encode fully mapped reactant–product pairs for each elementary step \cite{daylightsmirks, joung2025electron}, while others provide reactants, products, and a separate arrow code \cite{tavakoli2024pmechdb}, introducing redundancy. Formats that map only heavy atoms \cite{chen2024large} constrain how individual hydrogens can be referenced. Table \ref{tab: advantages_of_mechsmiles} provides a systematic comparison. Overall, MechSMILES balances expressiveness, human readability, and character efficiency in a way that is well-suited to the growing use of language models in chemistry \cite{white2023future, jablonka202314, bran2024transformers, ramos2025review, macknight2025rethinking}.

\begin{table}[h]
  \caption{Summary of advantages of MechSMILES for the encoding of elementary steps.\\flow: FlowER. m-us: mech-USPTO-31k, pmdb: PMechDB, msmi: MechSMILES. $^*$Results reported for the training split of PMechDB split-0, a visual representation is visible on Figure \ref{fig: char_efficiency_comparison}.}
  \label{tab: advantages_of_mechsmiles}
  \centering
  \begin{tabular}{lcccc}
    \hline
          &flow\citep{joung2025electron}& m-us \citep{chen2024large}& pmdb\citep{miller2025mechanism} & msmi (this work)\\
    \hline
    Arrows reported & \xmark & \cmark & \cmark & \cmark\\ 
    Reports stable intermediates & \cmark & \xmark & \cmark & \cmark\\
    Product implicit in the format & \xmark & \cmark & \xmark & \cmark\\
    Characters per elem. step $^*$ & 302 $\pm$ 223 & 176 $\pm$ 113 & 139 $\pm$ 65& \textbf{77 $\pm$ 33}\\

    \hline
  \end{tabular}
\end{table}

\subsection{Mechanistic step prediction models}
One of the most fundamental aspects of chemical transformations is the conservation of mass and charge. What occurs is a redistribution of electron density among the species involved. We designed our computational environment to encode this principle directly: models operating within it can only push arrows, making it impossible to create or destroy atoms or charges. This enforcement eliminates atom hallucination by construction. A constraint of this design is that all species contributing atoms or charges to the mechanism (catalysts, acids, bases, etc.) must be present in the initial state, as the environment will not authorize the use of species that were not provided.

Within this environment, we define four mechanism prediction tasks of increasing complexity (Figure \ref{fig: main_fig_task}). The simplest is an annotation task in which all chemical information of an elementary step is provided and the model must identify the electron movements coresponding to that transformation. The most challenging mirrors the problem a chemist faces in practice: given only the available reactant species and the desired product, without stoichiometry or by-product information, predict the complete sequence of arrow-pushing moves. At inference, the model serves as a policy that can be coupled with beam search or any other search algorithm to navigate multi-step mechanisms.

To demonstrate that our framework is architecture-agnostic, we trained two families of models: a T5 encoder-decoder architecture \cite{raffel2020exploring} and a LLaMa decoder-only architecture \citep{touvron2023llama}, both using a custom MechSMILES tokenizer mentioned in the previous section. These two models have been chosen as representatives of the two main transformer-based \cite{vaswani2017attention} generative architectures. Unlike recent works that focus on forward prediction through mechanistic moves, our models specialize in mechanism prediction when the product is already known. This post-hoc framing has a practical advantage: it allows the model to focus on pathways leading to a specific (potentially minor) product, and it enables the three downstream applications we demonstrate: reaction validation, holistic atom-to-atom mapping, and catalyst-aware template extraction.

\section{Results}\label{sec: results}
We first benchmark model performance across tasks of increasing difficulty, then demonstrate three applications enabled by mechanism prediction: post-hoc validation of CASP proposals, holistic atom-to-atom mapping including hydrogens, and extraction of catalyst-aware reaction
templates.

\subsection{Evaluation on tasks with increasing difficulty}

{\setlength{\itemsep}{0pt}\setlength{\parsep}{0pt}
\begin{itemize}
    \item \textbf{Task 1 -- Elementary Step Prediction:} Given current molecules and next intermediate, predict the electron movements. This is essentially a transcription and annotation task, as all chemical information is provided.
    \item \textbf{Task 2 -- Equilibrated Reaction:} Given current molecules and all final products (including by-products, with stoichiometry), predict the next step of the mechanism. This requires chemical planning to identify the sequence of elementary steps.
    \item \textbf{Task 3 -- Reaction without By-products:} Given current molecules with stoichiometry and main product only, predict the mechanism. The model must infer which species are consumed and ignore spectators.
    \item \textbf{Task 4 -- Reaction without Stoichiometry:} Given available species (without 
stoichiometry) and main product only, predict the mechanism. This mirrors real human-level mechanism prediction scenarios where only reactants, conditions and desired product are known.
\end{itemize}}
\begin{figure}[ht!]
\centering
\includegraphics[width=0.9\linewidth]{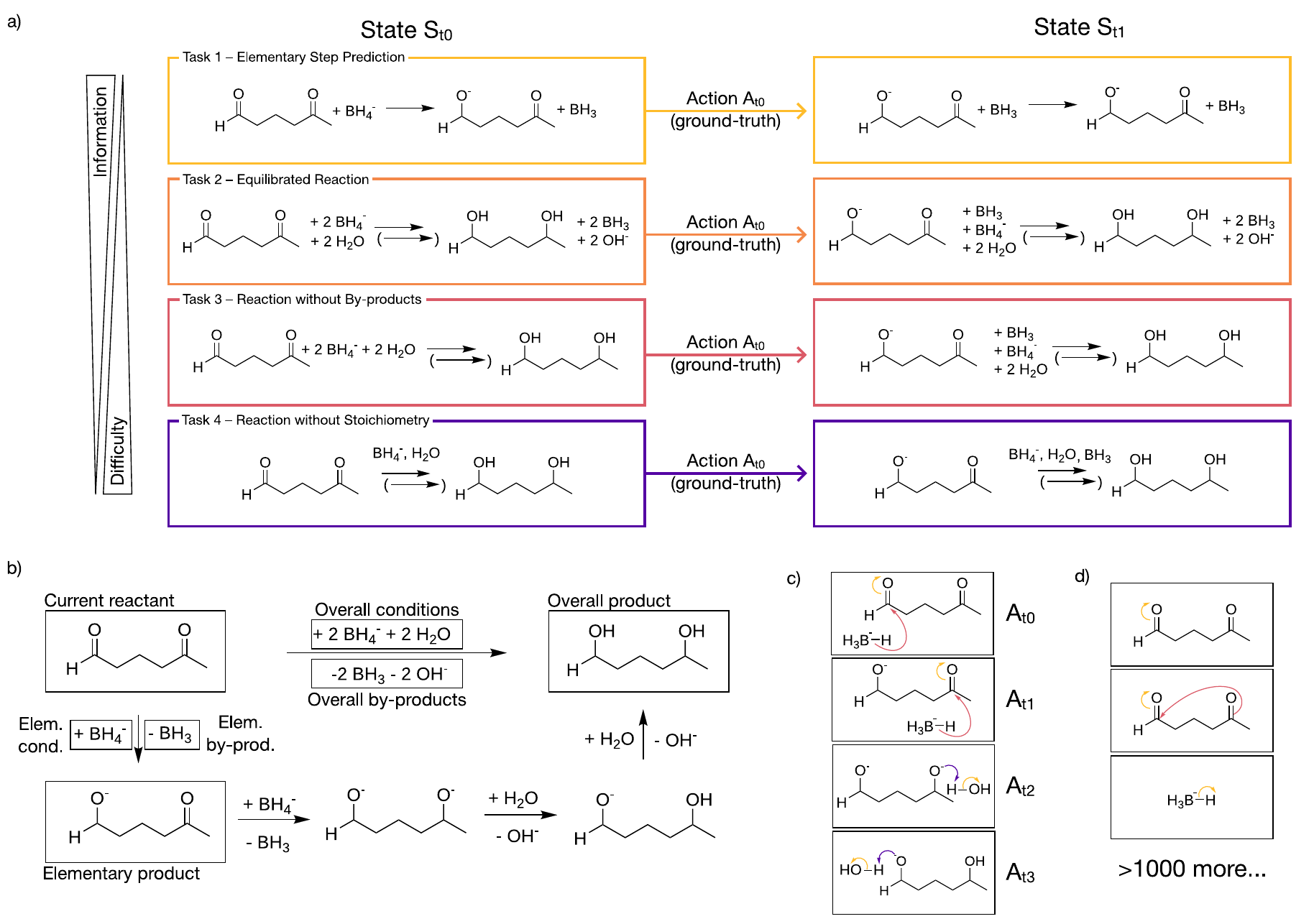}
\caption{Reaction mechanism prediction framework. (a) Progressive task difficulty showing elementary step prediction with decreasing input information, from Task 1 (full information) to Task 4 (no stoichiometry). Model receives state information (left boxes) and predicts actions as MechSMILES. (b) Complete reduction mechanism decomposed into elementary steps with reactants/conditions and products/by-products (distinction for visualization only). (c) Ground-truth action sequence from initial to goal state. (d) Sampled legal arrow pushing moves on state $S_{t0}$, where >1000 moves are possible but only a small subset are chemically meaningful and even fewer are productive.}
\label{fig: main_fig_task}
\end{figure}

We trained T5 encoder-decoder models \cite{raffel2020exploring} with a custom MechSMILES tokenizer and evaluated on two datasets: mech-USPTO-31k \cite{chen2024large} (random split at reaction level) and FlowER \cite{joung2025electron}  (original splits), with PMechDB \cite{tavakoli2024pmechdb} (original splits) used for the elementary step task. To test architecture generality, we also trained LLaMa decoder-only models \citep{touvron2023llama} on all tasks using the FlowER dataset. Table \ref{tab: main_perf_results} summarizes performance across all tasks and datasets. Details on training procedures and hyperparameters are available in the SI.

As expected, elementary step prediction achieves near-perfect accuracy on the two largest datasets (>96\% top-1 for both mech-USPTO-31k and FlowER), confirming that when all chemical information is provided, the model reliably annotates electron movements. This serves as a control experiment for the more demanding tasks. The underperformance on PMechDB likely reflects its substantially smaller size (8x smaller than mech-USPTO-31k and 110x smaller than FlowER in the number of elementary steps), which may be insufficient to train models of this scale.

For tasks requiring chemical planning (Tasks 2–4), top-1 step accuracy decreases modestly with increasing difficulty, though top-3 accuracy remains high, indicating that beam search consistently surfaces correct alternatives. On our most challenging task (no stoichiometry, no by-products), the T5 model achieves 95.7\% and 83.3\% top-1 step accuracy on mech-USPTO-31k and FlowER respectively. Complete mechanism retrieval reaches 73.3\% (mech-USPTO-31k) and 93.2\% (FlowER) with greedy decoding, rising to 86.5\% and 97.6\% with beam-width 3.

Two patterns in these results deserve comment. First, mech-USPTO-31k shows higher per-step accuracy than FlowER. This advantage likely reflects the random split, which allows training and test sets to share reaction template distributions. Second, pathway retrieval is higher than elementary step accuracy for FlowER, and lower for mech-USPTO-31k. These inverted trends are due to two competing effects. As a pathway is counted correct if all its elementary steps are correctly predicted, there is a compounding effect. With the length of a mechanism increasing, the probability of making an error is heightened. This effect explains why pathway retrieval can be lower than elementary step accuracy. On the other hand, for the elementary step accuracy metric, mechanisms are weighted proportionally to their length. This is not the case in the pathway retrieval metric, in which every whole mechanism is weighted equally. Since several mistakes in the same mechanism only count as one unrecovered pathway, long mechanisms don't impact this score as much. While the first effect seems to dominate for mech-USPTO-31k, the second one seems to dominate in FlowER. This can be understood by the proportion of very long mechanisms in the test set of FlowER compared to mech-USPTO-31k (Figure \ref{fig: test_set_length_distrib}).

\begin{table}[h!]
  \caption{Performance of models on mechanism prediction tasks ($^*$Results of FlowER models are reported for a related but different task: forward prediction via mechanisms. $^{**}$ Results reported from the original work). Elementary step accuracy measures correctness at each step; pathway retrieval measures the recovery of the complete mechanism for a given beam width Bw. n/a  indicates "not applicable", because the task has only one step (no complete mechanism to retrieve).\\flow: FlowER\citep{joung2025electron}, m-us: mech-USPTO-31k\citep{chen2024large}, pmdb: PMechDB split-0\citep{miller2025mechanism} }
  \label{tab: main_perf_results}
  \centering
\resizebox{\linewidth}{!}{\begin{tabular}{lllcc}
    \hline
    \multicolumn{3}{l}{\textbf{Training}} & \multicolumn{2}{l}{\textbf{Evaluation}}                   \\
    \cmidrule(r){1-5}
    Task     & Dataset & Model    &  Elem. step acc. [\%] & Pathway retrieval [\%]\\
    
             &             & architecture &  Top-1 (Top-3) & Bw = 1 (Bw = 3)\\
    \hline
    Elementary    &flow &      Encoder-decoder (T5)  &   99.70 (99.89)    & \multirow{4}{*}{n/a}  \\
    step &   & Decoder (LLaMa)      & 97.51 (98.65)  &   \\

                       &m-us  & Encoder-decoder (T5) &  96.63 (97.11)   &   \\
                       &pmdb & Encoder-decoder (T5) & 80.86 (85.15) & \\
    \cmidrule(r){2-5}

    Equilibrated             &flow &      Flow matching (FlowER$^*$)\citep{joung2025electron}  &   88.48  (97.96)$^{**}$    &    88.97 (96.48)$^{**}$   \\
    reaction&                   &      Flow matching (FlowER-large$^*$)\citep{joung2025electron}   &   89.74 (98.66)$^{**}$   &    92.50 (98.15)$^{**}$    \\ \cmidrule(r){3-5}
    &        flow           &     Encoder-decoder (T5)   &   84.27 (98.14)    &    96.02 (98.69)   \\
    &   & Decoder (LLaMa)       & 78.22 (92.63) &  76.92 (84.62)  \\
                 &m-us &  Encoder-decoder (T5) & 96.47 (97.35)    &   75.22 (87.37)    \\
    \cmidrule(r){2-5}
    
    Reaction           &flow  &      Encoder-decoder (T5)  &   84.21 (98.11)    &  95.86 (98.56)     \\
    (w/o by-prod.) &   & Decoder (LLaMa)       &  79.75 (93.69) & 74.26 (86.23)\\
       &m-us & Encoder-decoder (T5) &  96.48 (97.38)    &   76.03 (88.81)    \\
    \cmidrule(r){2-5}
    
    Reaction &flow         & Encoder-decoder (T5) &   83.33 (97.61)    &   93.16 (97.58)    \\
    w/o stoichio.&   & Decoder (LLaMa)       &   77.38 (91.83)& 68.24 (81.85)  \\
    (w/o by-prod.)&m-us & Encoder-decoder (T5) &  95.72 (96.56)    &  73.33 (86.54)   \\

    \hline
  \end{tabular}}
\end{table}

\subsection{Transfer learning to new classes (ozonolysis and Suzuki coupling)}
\begin{figure}[ht!]
\centering
\includegraphics[width=0.9\linewidth]{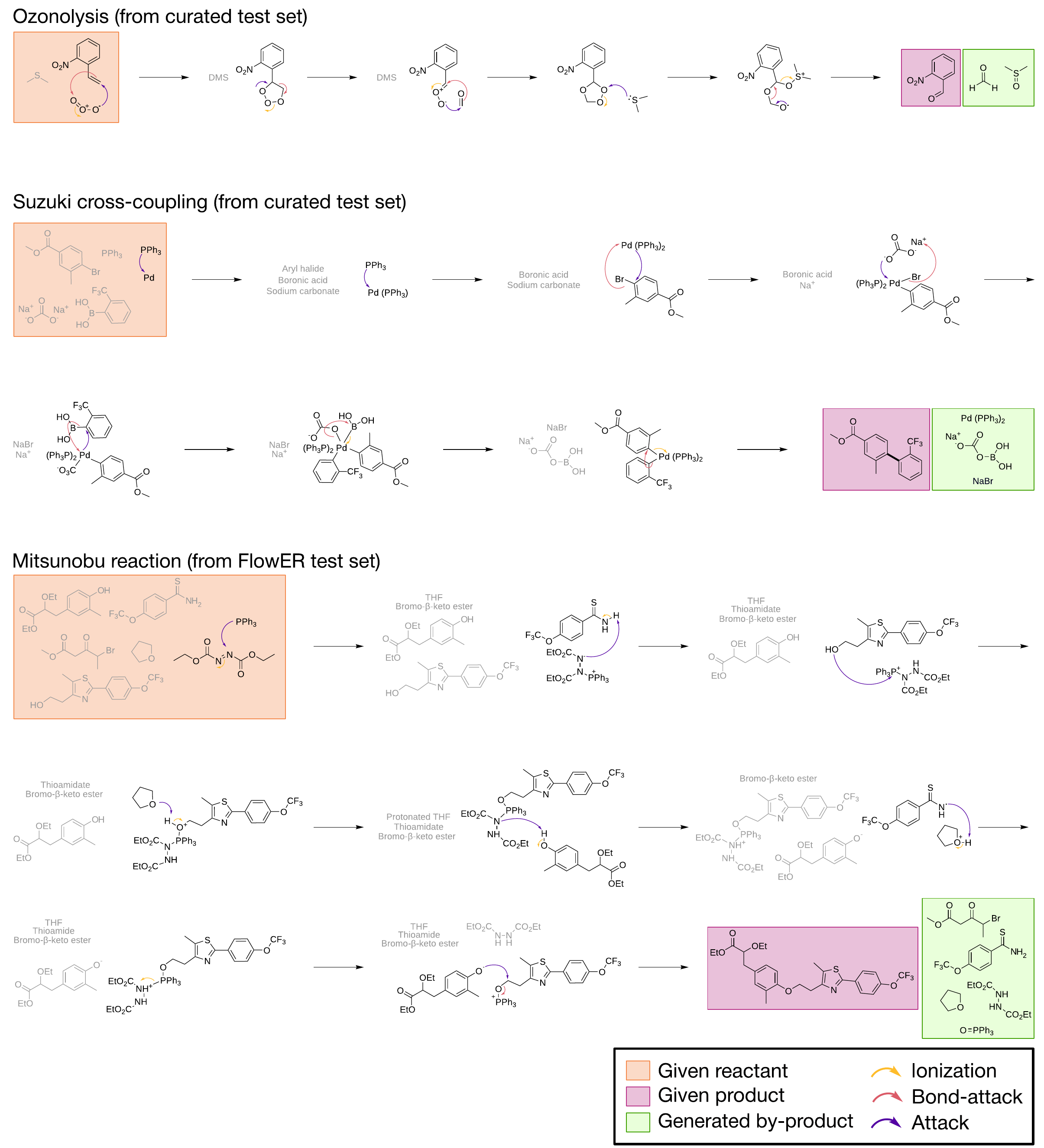}
\caption{Example test reactions that can be solved after fine-tuning on small curated datasets. While the base model (trained on the FlowER dataset without by-products task) achieves 0/5 and 1/8 accuracy on ozonolysis and Suzuki test sets respectively, the fine-tuned model trained on only 40 additional manually annotated examples for each class achieves to predict correctly 3/5 and 4/8 reactions in the test set, without forgetting complex mechanisms such as the Mitsunobu, that the model continues to predict correctly.}
\label{fig: transfer_learning}
\end{figure}

A practical question for any mechanism prediction framework is how quickly it can be extended to reaction classes absent from its training data. We tested this with two well-characterized transformations: ozonolysis with reductive workup and Suzuki cross-coupling with carbonate base. Both involve non-trivial multi-step mechanisms and were absent (ozonolysis) or severely underrepresented (11 examples for Suzuki) in the FlowER training set.

We curated small mechanistic datasets manually using a custom drag-and-drop annotation GUI (details in SI), selecting reactions of these two classes from USPTO-MIT 480k \cite{jin2017uspto_mit_480k}. The resulting datasets of 47 (ozonolysis) and 50 (Suzuki) annotated reactions were split into training, validation, and test sets (40/2/5 and 40/2/8 respectively). Baseline performances on the test reactions were 0\% (0/5) and 12.5\% (1/8) for ozonolyses and Suzukis respectively for the T5 model trained on task 3 on the FlowER dataset. After fine-tuning on the 40 training examples, performance rose to 60\% (3/5) for ozonolysis and 50\% (4/8) for Suzuki, with no degradation on previously learned mechanisms such as the Mitsunobu reaction (Figure \ref{fig: transfer_learning}). This result suggests that the framework has learned sufficiently general mechanistic reasoning to transfer to new reaction classes from as few as 40 manually annotated examples.

\subsection{Application 1: Explainability of a CASP model}
\begin{figure}[ht!]
\centering
\includegraphics[width=0.8\linewidth]{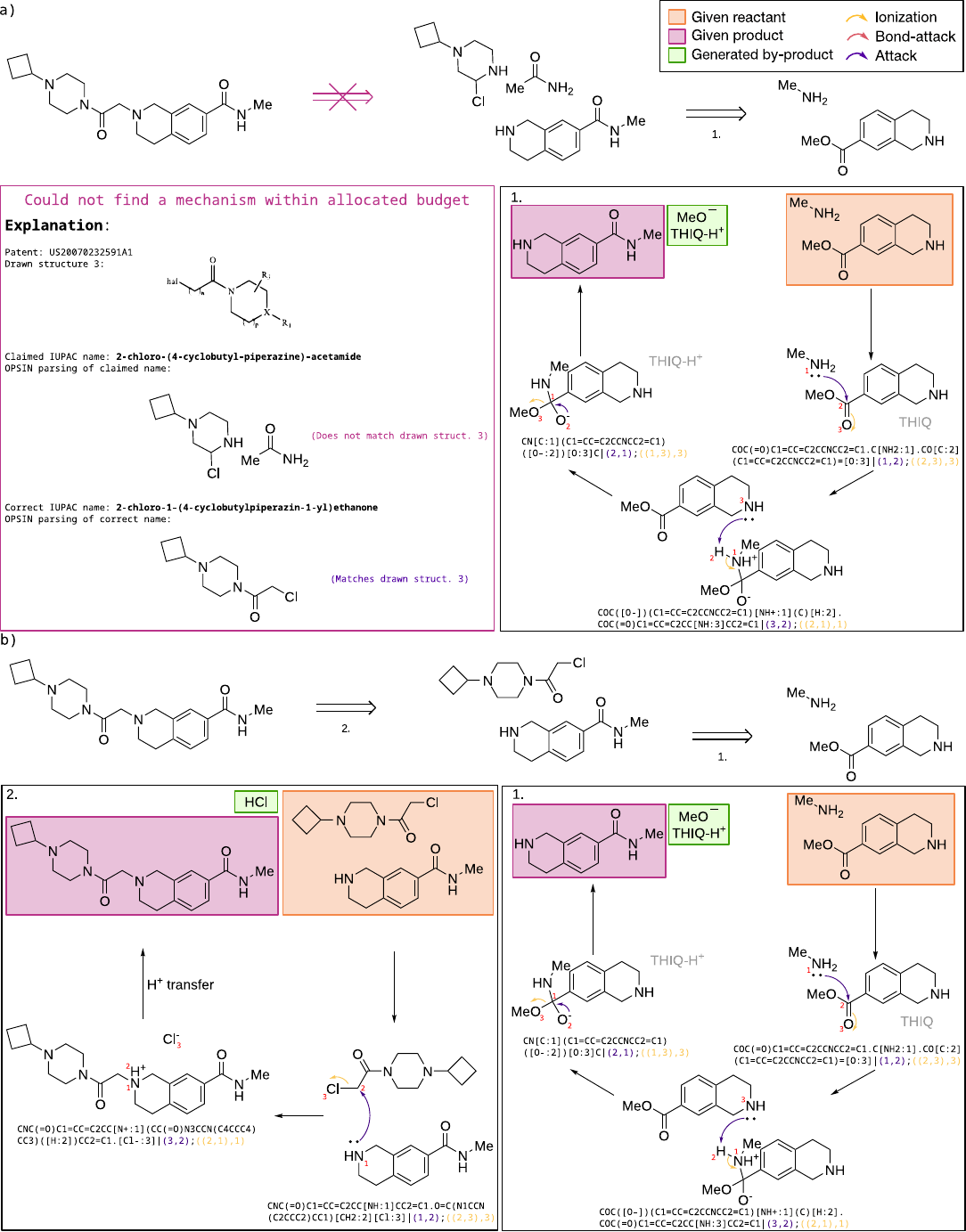}
\caption{a) Example of a CASP validation of the multistep reaction visible in Figure S2 of the PaRoutes paper\citep{genheden2022paroutes}. No mechanism can be found for the last reaction, with similar search settings as Figure \ref{fig: casp_validator_1} (same model, search algorithm and budget), hinting that this reaction might be wrong. After investigation, we found out this error might come from the name \textit{2-chloro-(4-cyclobutyl-piperazine)-acetamide} not respecting IUPAC rules in the original patent, potentially confusing tools such as OPSIN \cite{lowe2011chemical} for the creation of the molecules. b) The multistep reaction found in the original patent. After our correction, the now corrected last step finds a simple mechanism, hinting at the fact that this is indeed the correct transformation.}
\label{fig: casp_validator_2}
\end{figure}

A trained mechanism prediction model can validate individual reaction steps proposed by any CASP system, whether template-based or template-free. The validation workflow is straightforward: given a proposed reactant–product transformation, the model searches for a physically plausible sequence of arrow-pushing moves connecting them. If a mechanism is found, the transformation is supported; if not, it is flagged as potentially implausible.

We demonstrate this on a multistep retrosynthetic route from the PaRoutes benchmark \cite{genheden2022paroutes} (Figure \ref{fig: casp_validator_2}, with an additional example in Figure \ref{fig: casp_validator_1}). Our model successfully reconstructs mechanisms for all but one step. For the failing step, investigation revealed that the original patent contained a non-IUPAC chemical name that likely confused a tool such as OPSIN \cite{lowe2011chemical} into producing an incorrect molecular structure, and hence a transformation for which no valid mechanism exists. After correcting the structure to match the patent’s intended molecule, our model readily finds a simple mechanism, confirming the corrected transformation. This example illustrates how mechanistic validation can catch errors that propagate silently through existing CASP pipelines. For cases requiring additional confidence, the explicit electron pathway proposed by the model can be further assessed using computational energy calculations.

\subsection{Application 2: Holistic Atom-to-Atom Mapping Including Hydrogens}

Once a reaction mechanism is known, the role of every atom becomes transparent, enabling atom-to-atom mapping that tracks all species including hydrogens and by-products. This is a capability that current state-of-the-art mapping tools\cite{schwaller2021rxnmapper, chen2024localmapper} do not provide, primarily because they operate on implicit-hydrogen SMILES and compare initial and final states.

\begin{figure}[ht!]
\centering
\includegraphics[width=0.9\linewidth]{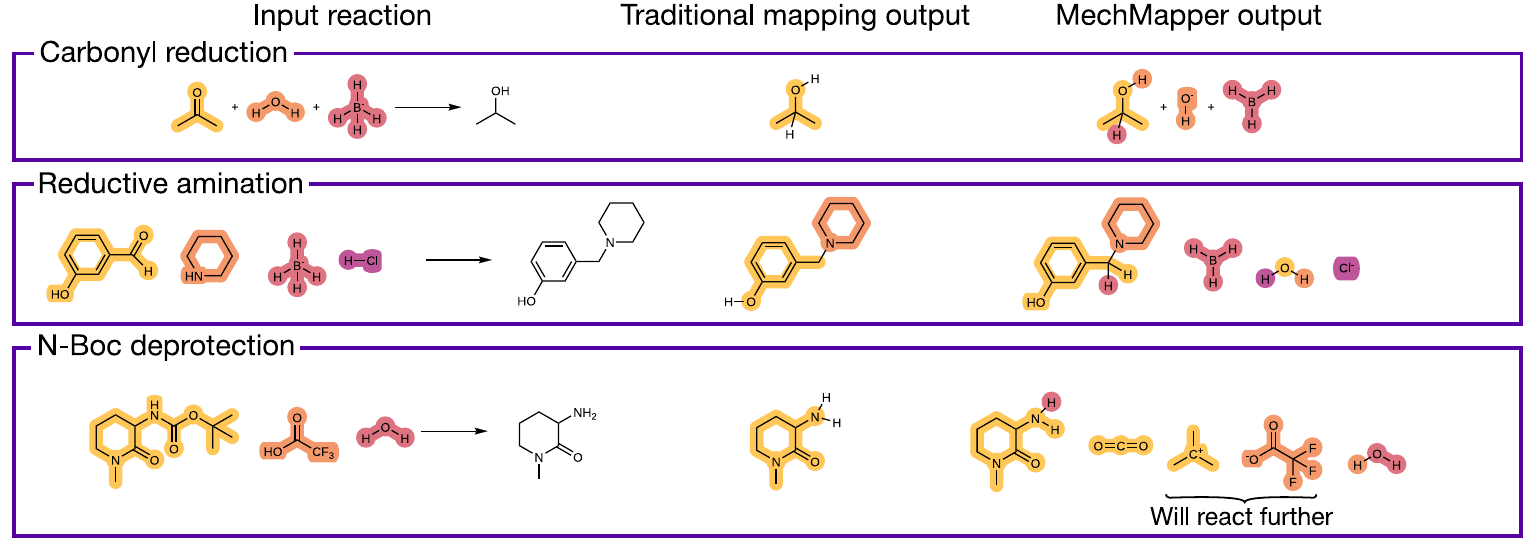}
\caption{Example reactions mapped both with SOTA tools, and with mechanistic mapping using our model as the mechanism predictor (MechMapper). Even though the mapping on heavy atoms is similar, the mechanism insight brings more information to the user, mapping hydrogens as well as by-products.}
\label{fig:h_aware_mapping}
\end{figure}

Figure \ref{fig:h_aware_mapping} illustrates this on three reactions: a carbonyl reduction by borohydride, a reductive amination, and an N-Boc deprotection. In each case, conventional mapping tools correctly assign heavy-atom correspondence but do not trace the origin of individual hydrogen atoms or predict by-products. Mechanistic mapping resolves both: in the carbonyl reduction, it identifies which hydrogens are transferred from borohydride versus water; in the reductive amination, it tracks the imine formation and reduction sequence; and in the N-Boc deprotection, it reveals that the carbamic acid undergoes a proton transfer with water. These examples span common synthetic transformations where hydrogen tracking is chemically consequential.

\subsection{Application 3: Catalyst-Aware Reaction Template Extraction}

\begin{figure}[ht!]
\centering
\includegraphics[width=0.9\linewidth]{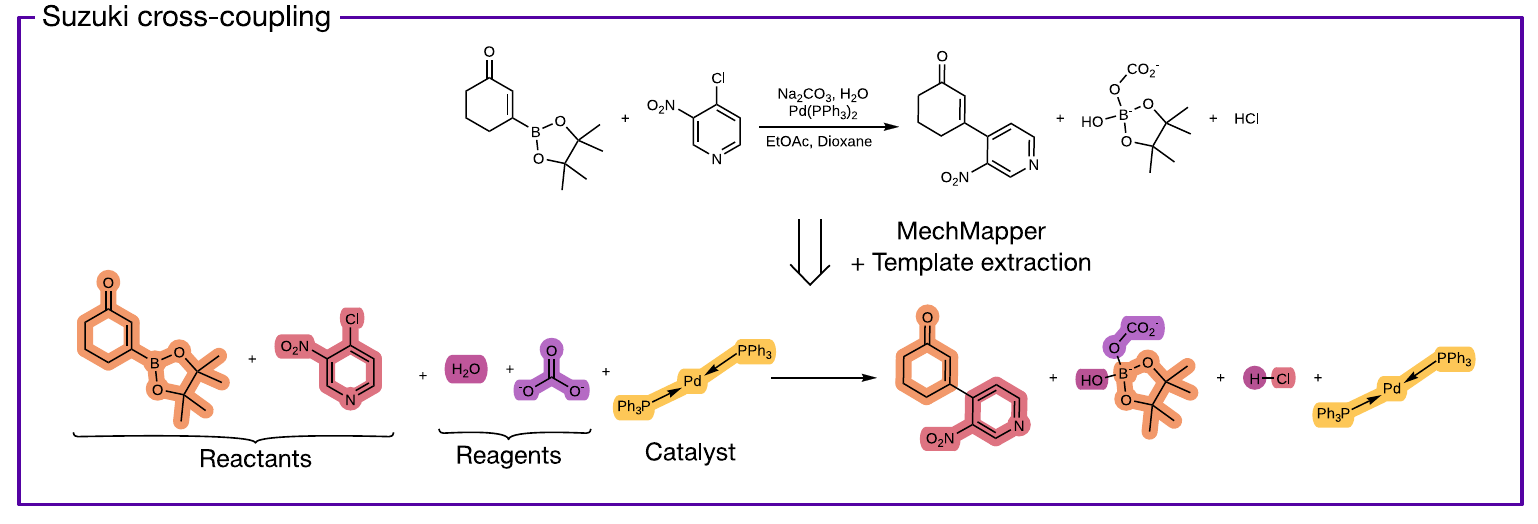}
\caption{On top, a Suzuki coupling reaction taken from the test set of the FlowER dataset\citep{joung2025electron}. \ce{Pd(PPh3)2} that can be seen over the arrow is a recycled catalyst. In the bottom, a visual representation of the mechanistic template of radius $=\infty$ of this reaction. The mechanistic template distinguishes a recycled catalyst from other pure observer species, and includes it in the template.}
\label{fig: cat_aware_template}
\end{figure}

Traditional reaction template creation, whether performed by experts \cite{synthia, joung2025electron} or via automatic SMARTS generation from atom-to-atom mapping \cite{kannas2022rxnutils}, generally compares only the initial and final states of a reaction. This makes it fundamentally blind to species that participate in intermediate steps but are regenerated by the end of the mechanism. Catalysts are the most important class of such species: they are often essential for the reaction to proceed, yet they are "invisible" in the net transformation and therefore absent from conventionally extracted templates.

Mechanistic understanding resolves this directly. Because our framework tracks every elementary step, it naturally distinguishes species that participate in intermediate stages (catalysts, certain acids and bases) from purely spectating species (solvents). Figure \ref{fig: cat_aware_template} illustrates this with a Suzuki cross-coupling from the FlowER test set: the palladium catalyst undergoes oxidative addition and reductive elimination before being regenerated. A conventional template derived from the net transformation would omit the palladium species entirely; our mechanistic template includes it. This principle applies broadly to any catalytic cycle and could serve as an automatic role-assignment method, complementing or replacing existing classification tools \cite{schneider2016roleassignment}.

\section{Discussion}
The results presented here establish that language models can learn to predict reaction mechanisms through arrow-pushing formalism with sufficient accuracy to be practically useful. On our most challenging task, which mirrors the information a chemist typically has available (reactants, conditions, and desired product), the framework retrieves the correct mechanism for the majority of test reactions across two independent datasets. Three implications follow.

First, mechanism prediction is ready to serve as a validation layer for existing CASP systems. Current synthesis planning tools, whether template-based or template-free, propose single-step transformations without mechanistic decomposition. Our framework can assess these proposals by searching for a physically plausible electron pathway, flagging transformations for which no mechanism can be found. As demonstrated with the PaRoutes example (Figure \ref{fig: casp_validator_2}), this approach can catch errors that propagate silently through retrosynthetic pipelines. Practical integration with CASP systems that do not explicitly propose reaction conditions would benefit from pairing our model with a condition prediction module.
Second, mechanism prediction unlocks downstream capabilities that operate at a level of granularity inaccessible to methods based on net transformations. Holistic atom-to-atom mapping, including hydrogen tracking, provides information critical for reactions where proton or hydride transfer play an important role. Catalyst-aware template extraction distinguishes species that participate in intermediate steps from true spectators, capturing mechanistic roles that are invisible when comparing only initial and final states. Both capabilities emerge naturally from the mechanistic representation and require no additional training or specialized models.
Third, the transfer learning results suggest that the framework can be extended to new reaction classes with minimal data curation effort. The ability to learn ozonolysis and Suzuki mechanisms from 40 annotated examples each, without degrading performance on previously learned classes, points toward a practical workflow: identify underperforming reaction classes, curate a small number of high-quality mechanistic annotations, and fine-tune. The custom annotation Graphical User Interface (GUI) we developed lowers the barrier to this curation step. Over time, this cycle could substantially expand the mechanistic coverage of the models.

Several limitations should be noted. The current framework is restricted to polar mechanisms involving closed-shell electron movements. While these represent the majority of synthetic organic transformations, extension to radical pathways would be necessary to cover the full landscape of organic reactivity. Additionally, our environment requires all species contributing atoms or charges to be present in the initial state; it cannot infer missing reagents. The evaluation datasets, while the largest available, are skewed toward common reaction classes, and performance on rare or highly complex mechanisms remains to be systematically assessed. Furthermore, while the reaction environment imposes no inherent constraint on stereochemistry, the lack of datasets annotated with stereo-informed mechanisms meant that stereochemical outcomes could not be evaluated within the current framework. Finally, although the framework is architecture-agnostic in principle, we have evaluated only two model families (T5 and LLaMa); benchmarking additional architectures would strengthen the generality claim.

Looking ahead, several directions follow naturally from this work. Extension to radical mechanisms would broaden applicability to photochemical reactions, and unlock some useful synthtic strategies such as radical brominations or Anti-Markovnikov additions. Integration of mechanism prediction with retrosynthetic search, rather than using it solely as a post-hoc filter, could enable CASP systems that plan at the mechanistic level, potentially discarding non-promising reactions faster. The low-data fine-tuning capability demonstrated here could be applied systematically across reaction classes to identify and fill gaps in mechanistic coverage. From a methodological perspective, the open-source environment and standardized benchmarking tasks provide a common framework for comparing future mechanism prediction systems, and we invite the community to contribute additional datasets and model architectures.

We see this work as a step towards more explainable and chemically grounded CASP, bridging part of the gap existing between black-box models and human chemists through a shared representation. The mechanistic formalism presented here provides a foundation for integrating different levels of abstraction, leading to a more holistic view of the synthetic process, and ultimately enabling more robust and interpretable tools for automated synthetic chemistry.

\section*{Acknowledgments}
The authors thank Robert S. Jordan (Intel) and Vijay Kris Narasimhan (Merck KGaA) for insightful discussions regarding the elaboration of this work.
T.A.N. acknowledges support from Intel and Merck KGaA via the AWASES programme. Z.J. acknowledges support by the Swiss National Science Foundation (SNSF) (grant number 214915). P.S. acknowledges support from the NCCR Catalysis (grant number 225147), a National Centre of Competence in Research funded by the Swiss National Science Foundation.
The authors acknowledge the use of LLMs, in particular Claude 4.6 for linguistic refinement and grammatical editing. All scientific content and interpretations remain the sole responsibility of the authors.

\section*{Data and code availability}
All datasets, models, and their tokenizers are available through the HuggingFace Hub, with all mechanistic steps being conveyed in the MechSMILES format. Code is available on GitHub. A live version of the drag and drop mechanism GUI is equally available at the time of submission.\\
HuggingFace Hub: \href{https://huggingface.co/collections/SchwallerGroup/chrimp-mechsmiles}{https://huggingface.co/collections/SchwallerGroup/chrimp-mechsmiles}\\
GitHub: \href{https://github.com/schwallergroup/ChRIMP}{https://github.com/schwallergroup/ChRIMP}\\
Graphical user interface: \href{https://theo-neu.com/mechanism_drag_n_drop}{https://theo-neu.com/mechanism\_drag\_n\_drop}

\nocite{namerxn}

\bibliography{biblio}

\appendix
\begin{suppinfo}
\setcounter{equation}{0}
\setcounter{figure}{0}
\setcounter{table}{0}
\renewcommand{\theequation}{Equ. S\arabic{equation}}
\renewcommand{\thefigure}{S\arabic{figure}}
\renewcommand{\thetable}{S\arabic{table}}

\section{Additional concrete example of CASP validation via mechanism prediction}

\begin{figure}[ht!]
\centering
\includegraphics[width=0.95\linewidth]{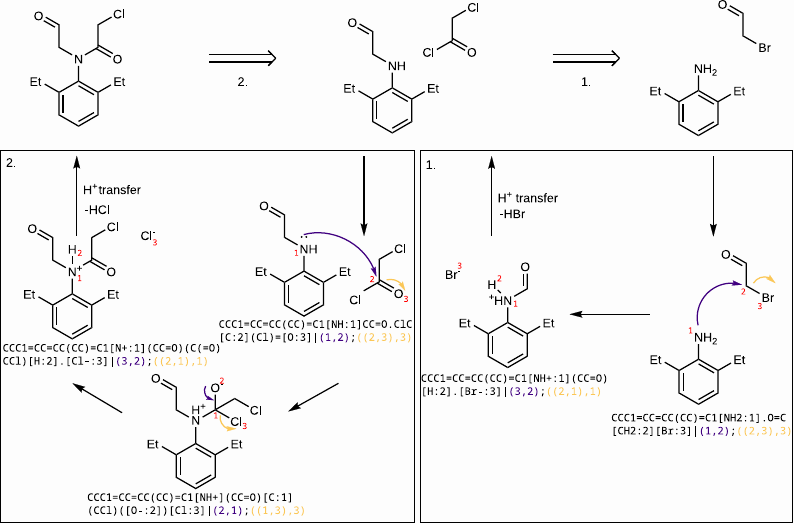}
\caption{Example of a CASP validation of the multistep reaction visible in figure 1 of the PaRoutes paper\citep{genheden2022paroutes}. Each step of this retrosynthesis (numbered arrows) is solved mechanistically in the frame below it, and with similar number. The mechanism found for each step comforts us in accepting these transformations as valid. All MechSMILES of this figure have been found by our model trained on the FlowER dataset, on the task of reaction without by-products. For the search algorithm, we used a best-first search algorithm with a maximum budget of 50 nodes to expand. In this particular case, the model performed perfectly, solving the arrow 1 by expanding 2 nodes, and arrow 2 by expanding 3 nodes.}
\label{fig: casp_validator_1}
\end{figure}

\section{Training format}
As mentioned in the main text, the performances exhibited in this article come uniquely from language-based models. Therefore we decided to train these models with special tokens telling the model which part of the input represents the reactants \texttt{[reac]}, and which part represents the product \texttt{[prod]}. Because we wanted to be able to train decoder-only models as well, we equally ended every input with the token \texttt{[mech]}, marking the end of the input and the beginning of the mechanism prediction. In order to let the model focus on the chemistry and not on positional artifacts, each training input has been shown in the forward direction (reactant before product) and in the retro direction (the opposite). Equally, to try to infuse the reactivity of species, training samples have  been shown to our models without the product. Despite the high performances observed with this methodology, an ablation study would be necessary to assess the utility of each of these data augmentations.\\

The inputs for the example task, shown in figure \ref{fig: main_fig_task}, as well as the expected output for all of them, are the following:

\textbf{Elementary steps}:\\
\breaktt{[prod]B.CC(=O)CCCC[O-][reac]CC(=O)CCCC=O.[BH4-][mech]}\\
\breaktt{[reac]CC(=O)CCCC=O.[BH4-][prod]B.CC(=O)CCCC[O-][mech]}\\
\breaktt{[reac]CC(=O)CCCC=O.[BH4-][mech]}

\textbf{Equilibrated reactions}:\\
\breaktt{[prod]B.B.CC(=O)CCCC[O-].[OH-].[OH-][reac]CC(=O)CCCC=O.O.O.[BH4-].[BH4-][mech]}\\
\breaktt{[reac]CC(=O)CCCC=O.O.O.[BH4-].[BH4-][prod]B.B.CC(=O)CCCC[O-].[OH-].[OH-][mech]}\\
\breaktt{[reac]CC(=O)CCCC=O.O.O.[BH4-].[BH4-][mech]}

\textbf{Stoichio. reactants, no by-prod}:\\
\breaktt{[prod]CC(=O)CCCC[O-][reac]CC(=O)CCCC=O.O.O.[BH4-].[BH4-][mech]}\\
\breaktt{[reac]CC(=O)CCCC=O.O.O.[BH4-].[BH4-][prod]CC(=O)CCCC[O-][mech]}\\
\breaktt{[reac]CC(=O)CCCC=O.O.O.[BH4-].[BH4-][mech]}

\textbf{No stoichio., no by-prod.}:\\
\breaktt{[prod]CC(=O)CCCC[O-][reac]CC(=O)CCCC=O.O.[BH4-][mech]}\\
\breaktt{[reac]CC(=O)CCCC=O.O.[BH4-][prod]CC(=O)CCCC[O-][mech]}\\
\breaktt{[reac]CC(=O)CCCC=O.O.[BH4-][mech]}

\textbf{Ground-truth output (minimal MechSMILES of the next arrow pushing move)}:\\
\texttt{[BH3-:2][H:3].[CH:4](CCCC(C)=O)=[O:1]|((2, 3), 4);((4, 1), 1)}

\section{Data}
\subsection{Statistics}
The work done in this article was enabled in parts thanks to the two large mechanistic datasets released by \citet{joung2025electron} and \citet{chen2024large}. The first has been released as a large dataset of mapped reactants and products of elementary steps, along with an index indicating which elementary steps are part of the same reaction. An algorithm was written to convert these mapped elementary reactions to MechSMILES, and the same splits as the original work were used for the training, validation and test sets respectively. The second has been released as multi-elementary steps directly, associated with all arrows necessary to pass from the initial reactants to final products. As no indication of the stable intermediates were given, we designed an algorithm checking for compliance with our environment for each arrow. Each intermediate that could be created in our environment has been deemed a stable intermediate of the mechanism. As no splits were explicitly given for this dataset, we performed a random split at the reaction level (keeping all elementary steps of the same reaction in the same split). Moreover, we also performed elementary step predictions on one split of the PMechDB dataset\cite{tavakoli2024pmechdb} released in a recent work\cite{miller2025mechanism}, this dataset is shared as a list of independent elementary steps, shared as a SMIRKS accompanied by an \textit{arrow-code}, effectively encoding for the transformation, and the arrows respectively. All the data that was used for training and evaluation are available publicly through HuggingFace datasets.\\

\begin{table}[h]
  \caption{Additional metrics on datasets.$^*$ Extracted successfully the with the snippet mentioned in section below. $^{**}$ Trivial elementary steps where no transformation happens and duplicates are not accounted for.}
  \label{tab: add_metrics_datasets}
  \centering
  \begin{tabular}{llll}
    \toprule
         Dataset & mech-USPTO-31k \citep{chen2024large}& FlowER\citep{joung2025electron}&PMechDB split-0\citep{miller2025mechanism}\\
    \midrule
    Reactions reported & 31,199 & 289,024 & not reported\\
    Reactions extracted$^*$ & 31,199 (100\%)& 259,834 (89.9\%) & n/a \\
    & & \\
    Elem. steps reported$^{**}$& not reported & 1,748,823 & 13,104\\
    Elem. steps extracted$^*$ & 114,826 & 1,435,192 (82.1\%)& 12,951 (98.8\%)\\
    & & \\
    Train/val/test splits & 80/10/10 & 89/1/10 & 80/10/10\\
    \bottomrule
  \end{tabular}
\end{table}

\subsection{Length distribution in the test sets}

\begin{figure}[ht!]
\centering
\includegraphics[width=1\linewidth]{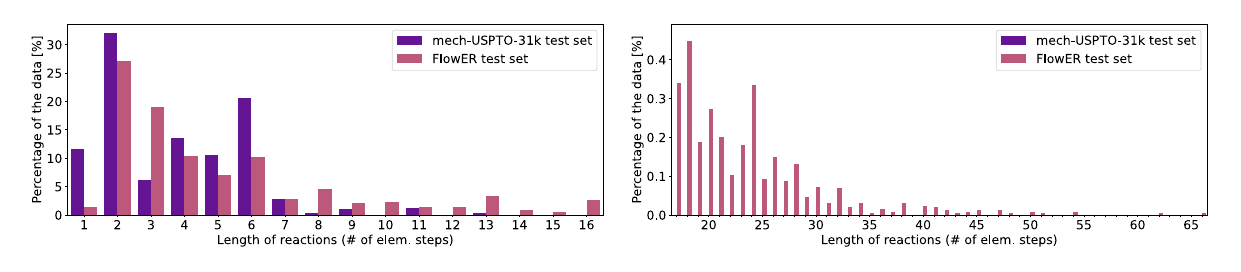}
\caption{Distribution of mechanism lengths (number of elementary steps) in the test sets of mech-USPTO-31k and FlowER.}
\label{fig: test_set_length_distrib}
\end{figure}

\subsection{Curation}
Snippets of code have been developed to translate the different formats of arrow pushing move to our MechSMILES format. In particular, we give access, along with our environment, to snippets of code capable of curating and translating the following formats:

\begin{enumerate}
\item Fully mapped elementary steps (similar to the Flower \citep{joung2025electron} format)
\item From a reactant and the set of all arrows (similar to the mech-USPTO-31k \citep{chen2024large} format)
\item SMIRKS accompanied by an arrow-code (similar to PMechDB \citep{tavakoli2024pmechdb} format)
\end{enumerate}

\begin{figure}[ht!]
\centering
\includegraphics[width=1\linewidth]{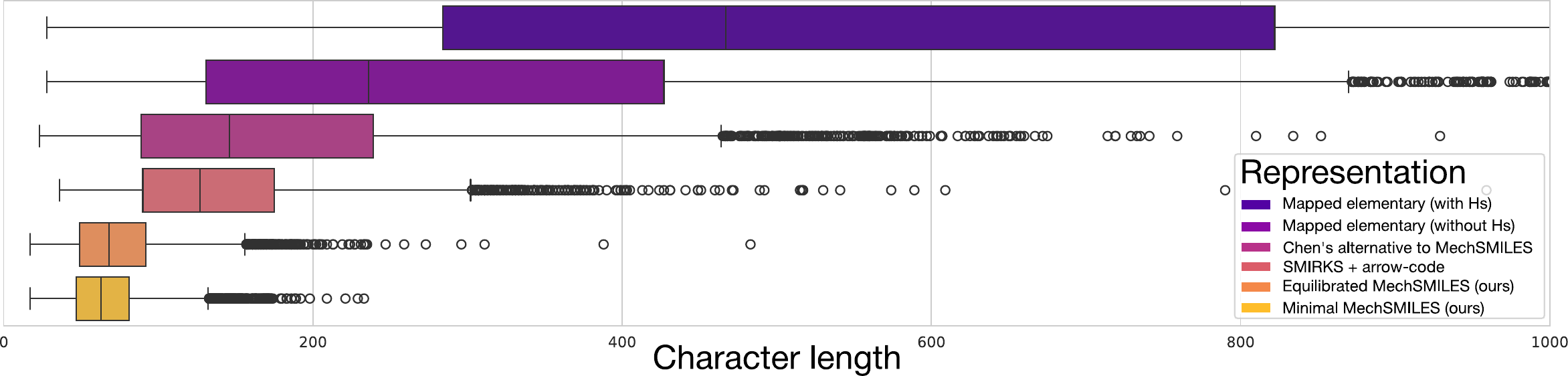}
\caption{Character length distribution to encode the same mechanistic data using the different formats mentioned in this work. The main difference between equilibrated and minimal MechSMILES is that the latter will not explicitely rewrite species that do not interact in the specific elementary step. The data used for this example is the training set of PMechDB split-0\citep{miller2025mechanism}}
\label{fig: char_efficiency_comparison}
\end{figure}

\section{Data curation for transfer learning}
\begin{figure}[ht!]
\centering
\includegraphics[width=1\linewidth]{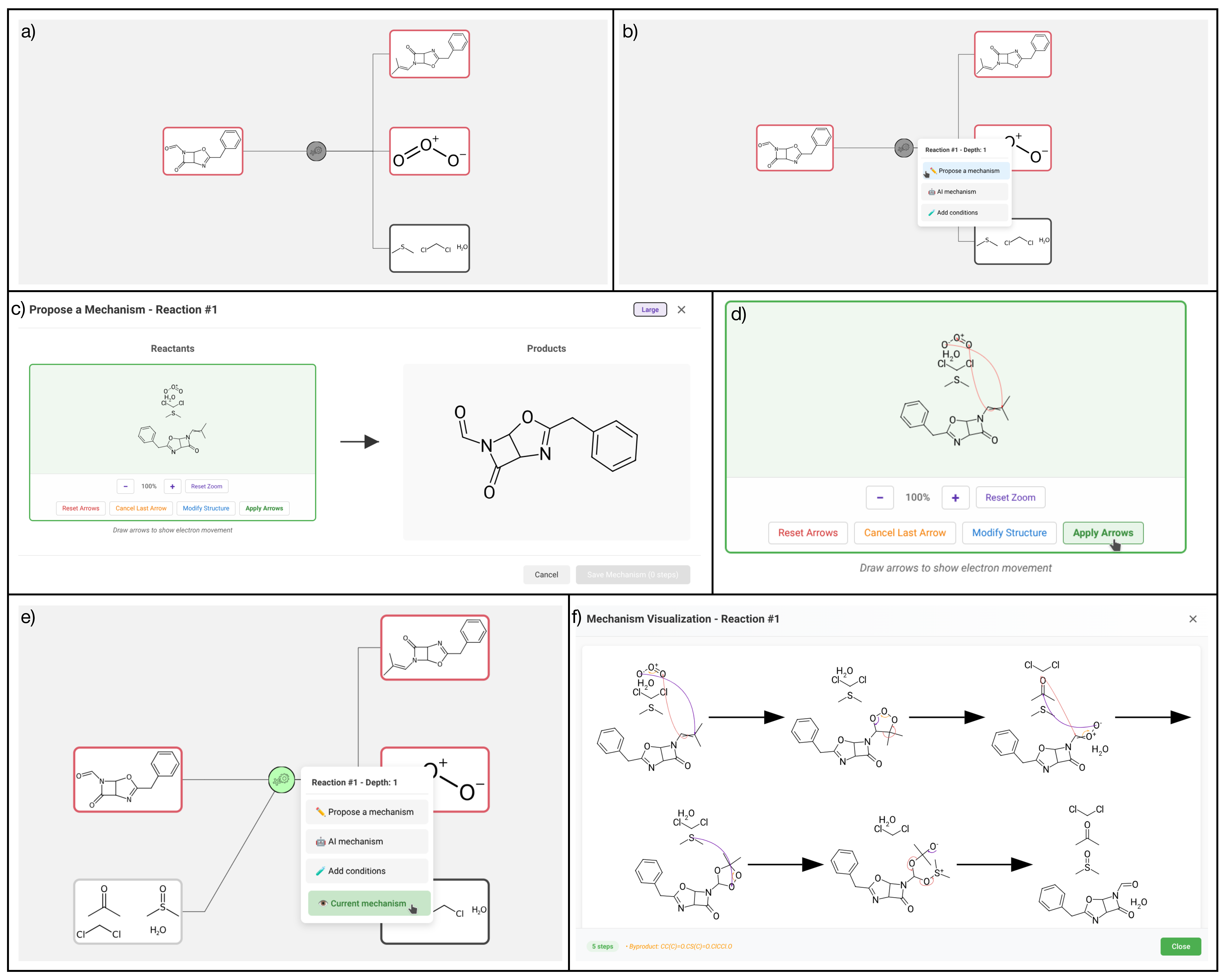}
\caption{Screenshot of the GUI used to create mechanistic database. a) Global view of the different nodes that form the product (right to left). b) The user can propose a mechanism manually. c-d) The manual mechanism panel opens, and the user can draw arrows by drag-and-drop on the structure, clicking "Apply arrows" to get the next intermediate, computed from the drawn arrows. e-f) Once a mechanism is informed for a reaction, the by-products are automatically computed and the user can get a global view to verify no mistake has been made. A \href{https://theo-neu.com/mechanism_drag_n_drop}{live demo of this tool} is available for test at the time of submission.}
\label{fig: mech_drag_n_drop_gui}
\end{figure}

To train a model on the transfer learning task, we first analyzed the USPTO-MIT 480K dataset \citep{jin2017uspto_mit_480k} and tagged reactions using NameRXN software \citep{namerxn}. For all classes (down to 2nd hierarchy level) containing at least 200 reactions (65 classes total), we randomly selected 20 representatives and attempted to predict a complete mechanism using our best T5 model on the "reaction without by-products" task, conducting 1,300 searches total. The parameters of this search were: budget of 50 nodes to expand, using a best-first search algorithm (using a heuristic score based of the sum of log-probabilities given by our model during inference), with maximum expansion of 5 nodes from the top-10 model predictions.\\
Following this screening, we performed manual inspection of the reaction classes and then selected two reaction classes that were absent or severely underrepresented in training and showed poor search performance: ozonolysis (0 training examples, 0/20 mechanisms solved) and Suzuki cross-coupling (11 training examples, 4/20 mechanisms solved). Note that these success rates represent a pessimistic lower bound on model capabilities, as many reactions in USPTO-MIT 480K lack reagents or conditions necessary for complete mechanistic prediction via arrow pushing.\\
To create high-quality training data for these classes, we manually curated two datasets using a custom web-based GUI with drag-and-drop arrow pushing capabilities (front-end) connected to our computational environment (back-end). Screenshots of this GUI are shown in Figure \ref{fig: mech_drag_n_drop_gui}. The final curated datasets consisted of 47 ozonolysis reactions with reductive workup and 50 Suzuki cross-coupling reactions with carbonate salt as base.

\section{Other architectures}
To display the capability of this format to be learnt by decoder-only models as well, we trained 4 models based on the LLaMa architecture on our 4 main tasks. These trainings have been performed on the FlowER dataset, being the biggest available, as a proof of concept that the environment could be used to train decoder-only architecture.

\section{Hyperparameters of model}
T5-model has been created with the default hyperparameters of HuggingFace's transformers package, except for the following list:

\begin{table}[h]
  \caption{Hyperparameters of T5-based model reported in the article}
  \label{tab: hyperparams_t5}
  \centering
  \begin{tabular}{ll}
    \toprule
    Key     & Value\\
    \midrule
    \texttt{d\_model} & 512\\
    \texttt{d\_ff} & 1024\\
    \texttt{num\_heads} & 6\\
    \texttt{num\_layers} & 4\\
    \texttt{vocab\_size} & tokenizer's vocabulary size (260)\\
    
    \bottomrule
  \end{tabular}
\end{table}

\begin{table}[h]
  \caption{Hyperparameters of LLaMa-based model reported in the article}
  \label{tab: hyperparams_llama}
  \centering
  \begin{tabular}{ll}
    \toprule
    Key     & Value\\
    \midrule
    \texttt{hidden\_size} & 512\\
    \texttt{intermediate\_size} & 512\\
    \texttt{num\_attention\_heads} & 4\\
    \texttt{num\_hidden\_layers} & 6\\
    \texttt{rope\_theta} & 8192\\
    \texttt{vocab\_size} & tokenizer's vocabulary size (260)\\
    \bottomrule
  \end{tabular}
\end{table}
\end{suppinfo}

\end{document}